# An Analysis of Phase Transition in NK Landscapes

**Yong Gao**                                           YGAO@CS.UALBERTA.CA
**Joseph Culberson**                                   JOE@CS.UALBERTA.CA
*Department of Computing Science*
*University of Alberta*
*Edmonton, Alberta, Canada, T6G 2H1*

## Abstract

In this paper, we analyze the decision version of the NK landscape model from the perspective of threshold phenomena and phase transitions under two random distributions, the uniform probability model and the fixed ratio model. For the uniform probability model, we prove that the phase transition is easy in the sense that there is a polynomial algorithm that can solve a random instance of the problem with the probability asymptotic to 1 as the problem size tends to infinity. For the fixed ratio model, we establish several upper bounds for the solubility threshold, and prove that random instances with parameters above these upper bounds can be solved polynomially. This, together with our empirical study for random instances generated below and in the phase transition region, suggests that the phase transition of the fixed ratio model is also easy.

## 1. Introduction

The *NK landscape* is a fitness landscape model devised by Kauffman (1989). An appealing property of the NK landscape is that the "ruggedness" of the landscape can be tuned by changing some parameters. Over the years, the NK landscape model itself has been studied from the perspectives of statistics and computational complexity (Weinberger, 1996; Wright, Thompson, & Zhang, 2000). In the study of genetic algorithms, NK landscape models have been used as a prototype and benchmark in the analysis of the performance of different genetic operators and the effects of different encoding methods on the algorithm's performance (Altenberg, 1997; Hordijk, 1997; Jones, 1995).

In the field of combinatorial search and optimization, one of the interesting discoveries is the threshold phenomena and phase transitions. Roughly speaking, a phase transition in combinatorial search refers to the phenomenon that the probability that a random instance of the problem has a solution drops abruptly from 1 to 0 as the order parameter of the random model crosses a critical value called the *threshold*. Closely related to this phase transition in solubility is the hardness of solving the problems. There has been strong empirical evidence and theoretical arguments showing that the hardest instances of the problems usually occur around the threshold and instances generated with parameters far away from the threshold are relatively easy. Since the seminal work of Cheeseman et al. (Cheeseman, Kanefsky, & Taylor, 1991), many NP-complete combinatorial search problems have been shown to have the phase transition and the associated easy-hard-easy pattern (Cook & Mitchell, 1997; Culberson & Gent, 2001; Freeman, 1996; Gent, MacIntyre, Prosser, &

Walsh, 1998; Kirkpatrick & Selman, 1994; Mitchell, Selman, & Levesque, 1992; Vandegriend & Culberson, 1998).

In this paper, we analyze the NK landscape model from the perspective of threshold phenomena and phase transitions. We establish two random models for the decision problem of NK landscapes and study the threshold phenomena and the associated hardness of the phase transitions in these two models.

The rest of the paper is organized as follows. In Section 2, we introduce the NK fitness landscape and our probabilistic models, the uniform probability model and the fixed ratio model. In Section 3 and Section 4, the threshold phenomena and phase transitions in NK landscapes are analyzed. For the uniform probability model, we prove that the phase transition of the uniform probability model is easy in the sense that there is a polynomial algorithm that can solve a random instance of the problem with the probability asymptotic to 1 as the problem size tends to infinity. For the fixed ratio model, we establish two upper bounds for the solubility threshold, and prove that random instances with parameters above these upper bounds can be solved polynomially. This, together with our empirical study for random instances generated below and in the phase transition region, suggests that the phase transition of the fixed ratio model is also easy. In Section 5, we report our experimental results on typical hardness of the fixed ratio model. In Section 6, we conclude our investigation and discuss implications of our results.

## 2. NK Landscapes and their Probabilistic Models

An NK landscape $f(x) = \sum_{i=1}^{n} f_i(x_i, \Pi(x_i))$, is a real-valued function defined on binary strings of fixed length, where $n > 0$ is a positive integer and $x = (x_1, \cdots, x_n) \in \{0, 1\}^n$. It is the sum of $n$ *local fitness functions* $f_i$, $1 \leq i \leq n$. Each local fitness function $f_i(x_i, \Pi(x_i))$ depends on the *main variable* $x_i$ and its neighborhood $\Pi(x_i) \subset \mathcal{P}_k(\{x_1, \cdots, x_n\} \backslash \{x_i\})$ where $\mathcal{P}_k(X)$ denotes the set of all subsets of size $k$ from $X$. The most important parameters of an NK landscape are the number of variables $n$, and the size of the neighborhood $k = |\Pi(x_i)|$.

In an NK landscape, the neighborhood $\Pi(x_i)$ can be chosen in two ways: the *random neighborhood*, where the $k$ variables are randomly chosen from the set $\{x_1, \cdots, x_n\} \backslash \{x_i\}$, and the *adjacent neighborhood*, where $k$ variables with indices nearest to $i$ (modulo $n$) are chosen. For example, for any even integer $k$, the $k$ variables in $\Pi(x_i)$ can be defined as $x_{((n+i-\frac{k}{2}) \mod n)}, \cdots, x_{((n+i+\frac{k}{2}) \mod n)}$. Once the variables in the neighborhood are determined, the local fitness function $f_i$ is determined by a fitness lookup table which specifies the function value $f_i$ for each of the $2^{k+1}$ possible assignments to the variables $x_i$ and $\Pi(x_i)$.

Throughout this paper, we consider NK landscapes with random neighborhoods. To simplify the discussion, we further assume that the local fitness functions take on binary values. Given an NK landscape $f$, the corresponding decision problem is stated as follows: Is the maximum of $f(x)$ equal to $n$? An NK landscape decision problem is insoluble if there is no solution for it.

It has been proved that the NK landscape model is NP complete for $k \geq 2$ (e.g., Weinberger, 1996; Wright et al., 2000). The proofs were based on a reduction from SAT to the decision problem of NK landscapes. To study the typical hardness of the NK landscape decision problems in the framework of thresholds and phase transitions, we introduce two random models. In both of the models defined below, the neighborhood set $\Pi(x_i)$ of a variable $x_i$ is selected by randomly choosing without replacement $k = |\Pi(x_i)|$ variables from $x \backslash \{x_i\}$.

**Definition 2.1.** *The Uniform Probability Model $\overline{N}(n, k, p)$:* In this model, the fitness value of the local fitness function $f_i(x_i, \Pi(x_i))$ is determined as follows: For each assignment $y \in Dom(f_i) = \{0, 1\}^{k+1}$, let $f_i(y) = 0$ with the probability $p$ and $f_i(y) = 1$ with the probability $1 - p$, where this is done for each possible assignment and each local fitness function independently.

**Definition 2.2.** *The Fixed Ratio Model $N(n, k, z)$:* In this model, the parameter $z$ takes on values from $[0, 2^{k+1}]$. If $z$ is an integer, we specify the local fitness function $f_i(x_i, \Pi(x_i))$ by randomly choosing without replacement $z$ tuples of possible assignments $Y = (y_1, \cdots, y_z)$ from $Dom(f_i) = \{0, 1\}^{k+1}$, and defining the local fitness function as follows:

$$f_i(y) = \begin{cases} 0, & if \ y \in Y; \\ 1, & else. \end{cases}$$

*For a non-integer $z = (1 - \alpha)[z] + \alpha[z + 1]$ where $[z]$ is the integer part of $z$, we choose randomly without replacement $[(1 - \alpha)n]$ local fitness functions and determine their fitness values according to $N(n, k, [z])$. The rest of the local fitness functions are determined according to $N(n, k, [z] + 1)$.*

In the theory of random graphs, there are two related random models $G(n, p)$ where each of the $\frac{n(n-1)}{2}$ possible edges is included in the graph independently with probability $p$, and $G(n, m)$ where exactly $m$ edges are chosen randomly and without replacement from the set of $\frac{n(n-1)}{2}$ possible edges. It is well known that for most of the monotone graph properties, results proved in $G(n, p)$ (or $G(n, m)$) also hold asymptotically for $G(n, Np)$ (correspondingly, $G(n, \frac{m}{N})$) where $N = \frac{n(n-1)}{2}$. However, we cannot expect that similar relations exist between the two random models of NK landscapes defined above unless the parameter $k$ tends to infinity. As a result, the asymptotic behaviors of the two NK landscape models are significantly different for fixed $k$.

We conclude this section by establishing a relation between the decision problem of NK landscapes and the SAT problem. A decision problem of the NK landscape

$$f(x) = \sum_{i=1}^{n} f_i(x_i, \Pi(x_i)),$$

"is the maximum of $f(x)$ equal or greater than $n$?", can be reduced to a (k+1)-SAT problem as follows:

(1) For each local fitness function $f_i(x_i, \Pi(x_i))$, construct a conjunction $C_i = \bigwedge_{j=1}^{z} C_i^j$ of clauses with exactly $k + 1$ variable-distinct literals from the set of variables $\{x_i, \Pi(x_i)\}$, where $z$ is the number of zero values that $f_i$ takes and $C_i^j$ is such that for any assignment $y_j \in \{0, 1\}^{k+1}$ that falsifies $C_i^j$, we have $f_i(y_j) = 0$.

(2) The (k+1)-SAT is the conjunction $\varphi = \bigwedge_{i=1}^{n} C_i$.

| $x$ | $y$ | $z$ | $f_i$ | Clauses |
|---|---|---|---|---|
| 0 | 0 | 0 | 0 | $x \vee y \vee z$ |
| 0 | 0 | 1 | 1 | |
| 0 | 1 | 0 | 1 | |
| 0 | 1 | 1 | 0 | $x \vee \bar{y} \vee \bar{z}$ |
| 1 | 0 | 0 | 1 | |
| 1 | 0 | 1 | 0 | $\bar{x} \vee y \vee \bar{z}$ |
| 1 | 1 | 0 | 0 | $\bar{x} \vee \bar{y} \vee z$ |
| 1 | 1 | 1 | 1 | |

Table 1: A local fitness function and its equivalent 3-clauses.

Table 1 shows an example of the fitness assignment of a local fitness function $f_i = f_i(x, y, z)$ and its associated equivalent 3-SAT clauses. It is easy to see that for any assignment $s$ to the variables $x, y, z$, $f_i(s) = 1$ if and only if the assignment satisfies the formula

$$x \vee y \vee z, \ x \vee \bar{y} \vee \bar{z}, \ \bar{x} \vee y \vee \bar{z}, \ \bar{x} \vee \bar{y} \vee z.$$

## 3. Analysis of The Uniform Probability Model

In the uniform probability model $\overline{N}(n, k, p)$, the parameter $p$ determines how many zero values a local fitness function can take. We are interested in how the solubility and hardness of the NK landscape decision problem change as the parameter $p$ increases from 0 to 1. It turns out that for fixed $p > 0$, the decision problem is asymptotically trivially insoluble. This is quite similar to the phenomena in the random models of the constraint satisfaction problem observed by Achlioptas et al. (1997).

To gain some more insight into the problem, we consider the case where $p = p(n)$ is a function of the problem size $n$ with $\lim_n p(n) = 0$. Our analysis shows that the solubility of the problem depends on how fast $p(n)$ decreases:

(1) If

$$\lim_n p(n) n^{\frac{1}{2^{k+1}}} = +\infty, \tag{3.1}$$

the problem is still asymptotically trivially insoluble because with the probability asymptotic to 1, there is at least one local fitness function that always has a fitness value 0;

(2) On the other hand if $p(n)$ decreases fast enough, i.e.,

$$\lim_n p(n) n^{\frac{1}{2^{k+1}}} < +\infty, \tag{3.2}$$

the problem can be decomposed into a set of independent sub-problems. In either case the problem can be solved in polynomial time. The case of (3.1) is not difficult to prove, but to prove the case of (3.2), we need to make use of the following concepts and results.

**Definition 3.1.** *The connection graph of an NK landscape instance* $f(x) = \sum\limits_{i=1}^{n} f_i(x_i, \Pi(x_i))$ *is a graph* $G = G(V, E)$ *satisfying*

*(1) Each vertex $v \in V$ corresponds to a local fitness function; and*

*(2) There is an edge between $v_i, v_j$ if and only if the corresponding local fitness functions $f_i, f_j$ share variables, i.e., the neighborhoods $\Pi(x_i)$ and $\Pi(x_j)$ of $x_i$ and $x_j$ have a non-empty intersection, and both of them have at least one zero value.*

**Definition 3.2.** *Let $f(x) = \sum_{i=1}^{n} f_i(x_i, \Pi(x_i))$ be an NK landscape instance with the connection graph $G = G(V, E)$. Let $G_1, \cdots, G_l$ be the connected components of $G$. Since the vertices of $G$ correspond to local fitness functions, we can regard $G_i$ as a set of local fitness functions. For each $1 \leq i \leq l$, let $U_i \subset x = (x_1, \cdots, x_n)$ be the set of variables that appear in the definition of the local fitness functions in $G_i$.*

It is easy to see that $(U_1, \cdots U_l)$ excluding independent vertices forms a disjoint partition of (a subset of) the variables $x = (x_1, \cdots, x_n)$, and that the local fitness functions in $G_i$ only depend on the variables in $U_i$. Furthermore, the NK decision problem is soluble if and only if for each $1 \leq i \leq l$, there is an assignment $s_i \in \{0, 1\}^{|U_i|}$ to the variables in $U_i$ such that for each local fitness function $g \in G_i$, $g(s) = 1$.

Theorem 3.1 summarizes the result on the uniform probability model.

**Theorem 3.1.** *For any $p(n)$ such that $\lim_n p(n) n^{\frac{1}{2^{k+1}}}$ exists, $k$ fixed, there is a polynomial time algorithm that successfully solves a random instance of $\overline{N}(n, k, p)$ with probability asymptotic to 1 as $n$ tends to infinity.*

Proof: We consider two cases: $\lim_n p(n) n^{\frac{1}{2^{k+1}}} = +\infty$ and $\lim_n p(n) n^{\frac{1}{2^{k+1}}} < +\infty$.

(1) The case of $\lim_n p(n) n^{\frac{1}{2^{k+1}}} = +\infty$.

Let $A_i$ be the event that $f_i(y) = 0$ for each possible assignment $y \in \{0, 1\}^{k+1}$ and let $A = \bigcup_{i=1}^{n} A_i$ be the event that at least one of the $A_i$'s occurs. We have

$$
\begin{aligned}
\lim_{n \to \infty} Pr\{A\} &= 1 - \lim_{n \to \infty} Pr\{\bigcap_{i=1}^{n} A_i^c\} \\
&= 1 - \lim_{n \to \infty} (1 - p(n)^{2^{k+1}})^n.
\end{aligned}
$$

It can be shown that if $k$ is fixed and $\lim_n p(n) n^{\frac{1}{2^{k+1}}} = +\infty$, then $\lim_{n \to \infty} Pr\{A\} = 1$. It follows that with probability asymptotic to one, there is at least one local fitness function which takes on values 0 for any possible assignments. We can therefore show that in this case, the NK decision problem is insoluble by checking the local fitness functions one by one. And this only takes linear time.

(2) The case of $\lim_n p(n) n^{\frac{1}{2^{k+1}}} < +\infty$.

Consider an algorithm that first finds the connected components $G_i$, $1 \leq i \leq l$ of the connection graph $G$ of the NK model, and then uses brute force to find an assignment $s_i \in \{0, 1\}^{|U_i|}$ to the variables in $U_i$ such that for each local fitness function $g \in G_i$, $g(s) = 1$. The time complexity of this algorithm is $O(n^2 + n * 2^{\mathcal{M}(n,k,p)})$ where $\mathcal{M}(n, k, p) = \max(|U_i|, 1 \leq i \leq l)$ is the maximum size of the subsets $(U_i, 1 \leq i \leq l)$ associated with the connected components of the connection graph. To prove the theorem, we only need to show that $\mathcal{M}(n, k, p) \in O(\log n)$. In the following, we will show that for $\lim_n p(n) n^{\frac{1}{2^k+1}} < +\infty$, we have

$$\lim_{n \to \infty} Pr\{\mathcal{M}(n, k, p) \leq 2^k + 2\} = 1$$

Consider the connection graph $G = G(V, E)$ of the NK model. It is a random graph and there is an edge between two nodes if and only if the two corresponding local fitness functions share variables and both of the local fitness functions take at least one zero as their fitness value. However, under this definition the edge probabilities are not independent. If $vx \in E$ then we know that $f_x$ has at least one zero and so the probability that $xw$ is in $E$ is greater than if there were no other edge on $x$.

To deal with this we resort to the following proof construction. Let $C_m = \{v_1, \ldots, v_m\}$ be a subset of $V$ of size $m$. Let $\pi$ be an ordering (permutation) of $v_1 \ldots v_m$. We say that $C_m$ is *variable connected with respect to the ordering* $\pi$, denoted as $\mathcal{C}(C_m, \pi)$, if for each $i, 2 \leq i \leq m$ there is either

1. a $j < i$ such that $f_{\pi(j)}$ and $f_{\pi(i)}$ share a variable; or

2. a $j, 1 \leq j \leq i$ such that the variable $x_j$ is one of the $k$ random variables in $f_i$.

**Lemma** *If the induced subgraph $G[C_m]$ is connected then there exists at least one ordering $\pi$ of $v_1 \ldots v_m$ such that $\mathcal{C}(C_m, \pi)$.*

As proof, consider the ordering of vertices of any depth first search of a connected subgraph. In this case, the connections are all by case 1.

The expected number of permutations $\pi$ for which $\mathcal{C}(C_m, \pi)$ is

$$E_c = \mathrm{E}[|\{\pi : \mathcal{C}(C_m, \pi)\}|] = m! \mathrm{Pr}\{\mathcal{C}(C_m, \pi)\}$$

We then observe that the expected number of connected induced graphs on $m$ vertices is less than $p_0^m \binom{n}{m} E_c$, where $p_0$ is the probability that $f_i$ takes at least one value zero. We show this value goes to zero in the limit if $m \geq 2^k + 2$. Finally, since if there is a connected subgraph on $m$ vertices then there must be one for each $i < m$, it follows that the largest connected component has size at most $2^k + 1$.

For a randomly generated permutation $\pi$ of $C_m$, let $C_i$ be the set of the first $i$ vertices of the permutation. For $i \geq 2$ define $P_i$ to be the probability that $f_{\pi(i)}$ shares at least one variable with $f_{\pi(j)}$ for some $j < i$ given that $\mathcal{C}(C_{i-1}, \pi/1, \cdots, i-1)$. Let $P_1 = 1$. (A one vertex subgraph is always connected.)

For $i > 1$ we have $P_i = \mathrm{Pr}\{\exists j < i, f_{\pi(i)} \text{ and } f_{\pi(j)} \text{ share variables, given } \mathcal{C}(C_{i-1}, \pi) \text{ or one of the } k \text{ random variables in } f_{\pi(i)} \text{ is in } \{x_1 \ldots x_m\} - \{x_i\}\}$.

$$\mathrm{Pr}\{\mathcal{C}(C_m, \pi)\} = \prod_{i=2}^{m} P_i.$$

Finally, for $i > 1$ we note that $C_{i-1}$ has at most $(i-1)k$ distinct other variables. If $C_{i-1}$ is connected then the number of variables may be less than this. Thus,

$$P_i \leq 1 - \frac{\binom{n-k(i-1)-m}{k}}{\binom{n-1}{k}}.$$

314

The combinatorial part reduces to

$$\frac{(n-k(i-1)-m)\ldots(n-k(i-1)-m-k+1)}{(n-1)\ldots(n-k)}$$
$$\geq \left(\frac{n-ki-m+1}{n-1}\right)^k.$$

So, $\Pr\{\mathcal{C}(C_m, \pi)\}$ is

$$\leq \prod_{i=2}^{m}\left(1-\left(\frac{n-ki-m+1}{n-1}\right)^k\right)$$
$$\leq \left(1-\left(1-\frac{km+m-2}{n-1}\right)^k\right)^{m-1}$$
$$\in O\left(\left(\frac{1}{n}\right)^{m-1}\right), m, k \text{ fixed.}$$

Noting that $p_0^m \in O\left(n^{\frac{-m}{2^k+1}}\right)$, we see that the expected number of connected subgraphs of size $m$ is bounded by

$$p_0^m\binom{n}{m}E_c \in O\left(n^m n^{\frac{-m}{2^k+1}}\left(\frac{1}{n}\right)^{m-1}\right)$$

which goes to zero if $m = 2^k+2$. It follows that $\mathcal{M}(n,k,p)$ is less than $2^k+2$ with probability asymptotic to 1. This completes the proof.

## 4. Analysis of The Fixed Ratio Model

As has been discussed in the previous section, the uniform probability model $\overline{N}(n,k,p)$ of NK landscapes is asymptotically trivial. Part of the cause of this asymptotic triviality lies in the fact that if the parameter $p$ does not decrease very quickly with $n$, then asymptotically there will be at least one local fitness function that takes the value 0 for all the possible assignments, making the whole decision problem insoluble. In this section, we study the fixed ratio model $N(n,k,z)$. In this model, we require that each local fitness function has fixed number of zero values so that the trivially insoluble situation in the uniform probability model is avoided. We note that the same idea has been used in the study of the *flawless CSP* (Gent et al., 1998).

Recall that in the fixed ratio model, we choose the neighborhood structure for each local fitness in the same way as in the uniform probability model $\overline{N}(n,k,p)$. To determine the fitness value for a local fitness function $f_i$, we randomly without replacement select exactly $z$ tuples $\{s_1, \cdots, s_z\}$ from $\{0,1\}^{k+1}$, and let $f_i(s_j) = 0$ for each $1 \leq j \leq z$ and $f_i(s) = 1$ for every other $s \in \{0,1\}^{k+1}$.

For the fixed ratio model, we are interested in how the probability of an instance of $N(n,k,z)$ being soluble changes as the parameter $z$ increases from 0 to $2^{k+1}$. It is easy to see that the property "There exists an assignment $x$ such that $f(x) = \sum_{i=1}^{n} f_i(x_i, \Pi(x_i)) = n$" is monotone in the parameter $z$ — the number of tuples at which a local fitness function takes zero. Actually, we have the following Lemma on the property of the solubility probability of the fixed ratio model:

**Lemma 4.1.** *For the fixed ratio model, if $z_1 > z_2$, then*

$$Pr\{N(n, k, z_1) \text{ is soluble}\} \leq Pr\{N(n, k, z_2) \text{ is soluble}\}.$$

*Furthermore, we have*

$$Pr\{N(n, k, z) \text{ is soluble}\} = \begin{cases} 1, & \text{if } z \leq 1; \\ 0, & \text{if } z = 2^{k+1}. \end{cases}$$

Based on the above Lemma and in parallel to the study of the threshold phenomena in other random combinatorial structures such as 3-Coloring of random graphs and random 3-SAT, we suggest the following conjecture:

**Conjecture 4.1.** *There exists a threshold $z_c$ such that*

$$\lim_{n \to \infty} Pr\{N(n, k, z) \text{ is soluble}\} = \begin{cases} 1, & \text{if } z < z_c; \\ 0, & \text{if } z > z_c. \end{cases}$$

Conjectures like this are the starting point of the study of phase transition in many random combinatorial structures such as 3-coloring of random graphs and random SAT, but the existence of the thresholds is still an open question (Achlioptas, 1999; Cook & Mitchell, 1997). However, bounding the thresholds has been an important topic in the study of phase transition (Achlioptas, 1999, 2001; Dubois, 2001; Franco & Gelder, 1998; Franco & Paul, 1983; Frieze & Suen, 1996; Kirousis, P.Kranakis, D.Krizanc, & Y.Stamation, 1994). In this section, we will establish two upper bounds on the threshold of the parameter $z_c$, and theoretically prove that random instances generated with the parameter $z$ above these upper bounds can be solved with probability asymptotic to 1 by polynomial (even linear) algorithms.

Characterizing the sharpness of the thresholds is also of great interest in the study of the phase transition. After proving that every monotone graph property has a threshold behavior (Friedgut & Kalai, 1996), Friedgut (1999) established a necessary and sufficient condition for a monotone graph property to have sharp threshold, which has been used to prove the sharpness of the thresholds of 3-colorability and 3-SAT problems (Friedgut, 1999; Achlioptas, 1999). For the fixed ratio model discussed in this paper, we suspect that it will exhibit a coarse threshold behavior, and would like to leave a detailed investigation into this problem as a future research direction.

## 4.1 The Upper Bound of $z = 3.0$

The derivation of this upper bound is based on the concept of a *conflicting pair* of local fitness functions. We say that two local fitness functions $f_i$ and $f_j$ conflict with each other if

1. $f_i$ and $f_j$ share at least one variable $x$; and

2. For any assignment $s \in \{0,1\}^n$, we have $f_i(s)f_j(s) = 0$.

It is obvious that an instance of the NK decision problem is insoluble if there exists a pair of conflicting local fitness functions.

Based on the second moment method in the theory of probability (Alon & Spencer, 1992), we can prove the following upper bound result. As it takes linear time to check if there is a pair of conflicting local fitness functions, we can see that the fixed ratio model $N(n,2,z)$ is linearly solvable when $z > 3.0$.

**Theorem 4.1.** *Define $A$ to be the event that there is a conflicting pair of local fitness functions in $N(n,2,z)$. For the fixed ratio model $N(n,2,z)$ with $z = 3.0 + \varepsilon$, we have*

$$\lim_n Pr\{A\} = 1$$

*and thus the problem is insoluble with probability asymptotic to 1.*

*Proof:* Without loss of generality, we may write $f$ as where $f_i$ has 4 zeroes in its fitness value assignment for $1 \leq i \leq \varepsilon n$, and 3 zeroes for $\varepsilon n + 1 \leq i \leq n$. Let $I_{ij}$ be the indicator function of the event that $f_i$ and $f_j$ conflicts with each other, i.e.,

$$I_{ij} = \begin{cases} 1, & \text{if } f_i \text{ and } f_j \text{ conflicts with each other;} \\ 0, & \text{else.} \end{cases}$$

and $S = \sum_{1 \leq i,j \leq \varepsilon n} I_{ij}$. We claim that $\lim_{n \to \infty} Pr\{S = 0\} = 0$.

By Chebyschev's inequality, we have

$$Pr\{S = 0\} \leq Pr\{|S - E(S)| \geq E(S)\}$$
$$\leq \frac{Var(S)}{(E(S)^2)}. \tag{4.3}$$

Since for each $1 \leq i \leq \varepsilon n, f_i$ has exactly 4 zeros in its fitness value assignment, we know that two local fitness function $f_i, f_j, 1 \leq i,j \leq \varepsilon n$, conflict with each other if and only if they have exactly one common variable $x$ such that one of the following is true: (1) $f_i(s) = 0$(or 1), $f_j(s) = 1$(or 0) for all the assignments $s$ such that $x = 1$(respectively $x = 0$); and (2) $f_i(s) = 1$(or 0), $f_j(s) = 0$(or 1) for all the assignments $s$ such that $x = 1$(respectively $x = 0$);

Since the probability that two local fitness functions share at least one variable is equal to

$$1 - \frac{\binom{n-2}{2}\binom{n-4}{2}}{\binom{n-1}{2}\binom{n-1}{2}},$$

we have

$$Pr\{I_{ij} = 1\} = \left(1 - \frac{\binom{n-2}{2}\binom{n-4}{2}}{\binom{n-1}{2}\binom{n-1}{2}}\right) \cdot 2\left(\frac{1}{\binom{8}{4}}\right)^2$$
$$= \Omega(\frac{1}{n}), \quad \varepsilon > 0, \quad 1 \leq i,j \leq \varepsilon n, \tag{4.4}$$

317

and hence,

$$E(S) = \sum_{1 \leq i,j \leq \varepsilon n} E(I_{ij}) = \sum_{1 \leq i,j \leq \varepsilon n} Pr\{I_{ij} = 1\} \in \Omega(n).$$

We now consider the variance of $S$. Since $S = \sum_{1 \leq i,j \leq \varepsilon n} I_{ij}$, we have

$$Var(S) = \frac{\sum_{i,j} Var(I_{ij}) + 2 \sum_{(i,j) \neq (l,m)} [E\{I_{ij}I_{lm}\} - E\{I_{ij}\}E\{I_{lm}\}]}{(E(S))^2}.$$

Let

$$A_1 = \frac{\sum_{i,j} Var(I_{ij})}{(E(S))^2}$$

and

$$A_2 = \frac{2 \sum_{(i,j) \neq (l,m)} [E\{I_{ij}I_{lm}\} - E\{I_{ij}\}E\{I_{lm}\}]}{(E(S))^2}.$$

It is easy to see that $\lim_{n \to \infty} A_1 = 0$. To prove $\lim_{n \to \infty} A_2 = 0$, we consider two cases:

**Case 1:** $i \neq j \neq m \neq l$. In this case, the two random variables $I_{ij}$ and $I_{lm}$ are actually independent. It follows that $E\{I_{ij}I_{lm}\} - E\{I_{ij}\}E\{I_{lm}\} = 0$.

**Case 2:** $(i,j) \neq (l,m)$, but they have one in common, say $j = l$. In this case, we have

$$E\{I_{ij}I_{lm}\} - E\{I_{ij}\}E\{I_{lm}\} = Pr\{I_{ij} = 1\}Pr\{I_{jm} = 1 | I_{ij} = 1\} - \Omega\left(\left(\frac{1}{n}\right)^2\right)$$

$$= \Omega\left(\frac{1}{n}\right)Pr\{I_{jm} = 1 | I_{ij} = 1\} - \Omega\left(\left(\frac{1}{n}\right)^2\right)$$

Given that $f_i$ and $f_j$ conflict with each other, the conditional probability that $f_j$ and $f_m$ conflict with each other is still in $\Omega(\frac{1}{n})$.

Since there are only $C_{\varepsilon n}^3$ pairs of $I_{ij}$ and $I_{jm}$ satisfying the condition in Case 2, we know that $\sum_{(i,j) \neq (l,m)} [E\{I_{ij}I_{lm}\} - E\{I_{ij}\}E\{I_{lm}\}]$ is in $\Omega(n)$. And therefore, $\lim_{n \to \infty} A_2 = 0$. It follows that

$$\lim_{n \to \infty} Pr\{S = 0\} \leq \lim_{n \to \infty} \frac{Var(S)}{(E(S)^2)} = 0.$$

Since the event $\{S > 0\}$ implies that there exists a conflicting pair of local fitness functions, the theorem follows. $\square$

## 4.2 2-SAT Sub-problems in $N(n, 2, z)$ and a Tighter Upper Bound

In this subsection, we establish a tighter upper bound $z > 2.837$ for the threshold of the fixed ratio model $N(n, 2, z)$ by showing that asymptotically $N(n, 2, z)$ contains an unsatisfiable 2-SAT sub-problem with probability 1 for any value of $z$ greater than 2.873. This also gives us a polynomial time algorithm which determines that $N(n, 2, z)$ is insoluble with probability asymptotic to 1 for $z > 2.837$.

Recall from Section 2 that each instance of $N(n, 2, z)$ has an equivalent 3-SAT instance. The idea is to show that with probability asymptotic to 1, an instance of $N(n, 2, z)$ will contain a set of specially structured 3-clauses, called a t-3-module (Definition 10.3, Franco & Gelder, 1998):

$$\mathcal{M} = \{M_1, \ldots, M_{3p+2}\}, \quad t = 3p + 2,$$

where

$$
\begin{aligned}
M_1 &= (\bar{u}_1 \vee u_2 \vee z_1, \bar{u}_1 \vee u_2 \vee \bar{z}_1); \\
&\cdots \\
M_{p-1} &= (\bar{u}_{p-1} \vee u_p \vee z_{p-1}, \bar{u}_{p-1} \vee u_p \vee \bar{z}_{p-1}); \\
M_p &= (\bar{u}_p \vee \bar{u}_0 \vee z_p, \bar{u}_p \vee \bar{u}_0 \vee \bar{z}_p); \\
M_{p+1} &= (\bar{u}_{p+1} \vee u_{p+2} \vee z_{p+1}, \bar{u}_{p+1} \vee u_{p+2} \vee \bar{z}_{p+1}); \\
&\cdots \\
M_{3p-1} &= (\bar{u}_{3p-1} \vee u_{3p} \vee z_{3p-1}, \bar{u}_{3p-1} \vee u_{3p} \vee \bar{z}_{3p-1}); \\
M_{3p} &= (\bar{u}_{3p} \vee u_0 \vee z_{3p}, \bar{u}_{3p} \vee u_0 \vee \bar{z}_{3p}) \\
M_{3p+1} &= (\bar{u}_0 \vee u_1 \vee z_{3p+1}, \bar{u}_0 \vee u_1 \vee \bar{z}_{3p+1}); \\
M_{3p+2} &= (u_0 \vee u_{p+1} \vee z_{3p+2}, u_0 \vee u_{p+1} \vee \bar{z}_{3p+2});
\end{aligned}
$$

and $u_1, \cdots, u_{3p+1}, z_1, \cdots, z_{3p+1}$ are binary variables. Notice that a t-3-module can be reduced to a 2-SAT problem containing two contradictory cycles and hence is unsatisfiable.

The result is proved in two steps. In the first step, it is shown that for $z > 2.837$ the average number of t-3-modules contained in $N(n, 2, z)$ tends to infinity as $n$ increases. In the second step, we use a result established by Alon and Spencer (1992) on the second moment method to prove that for $z > 2.837$ the probability that $N(n, 2, z)$ contains at least one t-3-module tends to 1.

Let us start with the first step to show that the average number of t-3-modules contained in $N(n, 2, z)$ tends to infinity as $n$ increases.

**Definition 4.1.** *Given a t-3-module $\mathcal{M}$ and an NK landscape instance $f = \sum_{i=1}^{n} f_i, k = 2$, a sequence of local fitness functions*

$$\mathbf{g} = (g_1, \cdots, g_t) \subset (f_1, \cdots, f_n)$$

*is said to be a possible match(PM) if for each $1 \le m \le t$, the main variable of $g_m$ is one of the three variables that occur in the 3-module $M_m$. A subsequence $(h_1, \cdots, h_l)$ of a possible match $\mathbf{g}$ is legal if for any $1 \le m < j \le l$, $h_m \neq h_j$.*

**Lemma 4.2.** *Let $f(x) = \sum_{i=1}^{n} f_i(x_i, \Pi(x_i))$ be an instance of $N(n, 2, z)$ and $\mathcal{M}$ be a t-3-module. Then the number of possible matches for the t-3-module $\mathcal{M}$ is $3^t$. Further, the number of legal possible matches is $\Theta\left(\left(\frac{3+\sqrt{5}}{2}\right)^t\right)$.*

*Proof:* For each $1 \leq m \leq t$, there are exactly 3 possible choices for $g_m$:

$$f_{i_1}(x_{i_1}, \Pi(x_{i_1})), f_{i_2}(x_{i_2}, \Pi(x_{i_2})), f_{i_3}(x_{i_3}, \Pi(x_{i_3})),$$

where $x_{i_1}, x_{i_2}, and \ \ x_{i_3}$ correspond to the three variables that occur in the 3-module $M_m$. Therefore, there are $3^t$ possible matches for the t-3-module.

To prove the second conclusion, we divide the t-3-module into 3 parts $\mathcal{M} = (\mathcal{M}_1, \mathcal{M}_2, \mathcal{M}_3)$, where $\mathcal{M}_1 = (M_m, 1 \leq m \leq p)$, $\mathcal{M}_2 = (M_m, p + 1 \leq m \leq 3p - 1)$, and $\mathcal{M}_3 = (M_{3p}, M_{3p+1}, M_{3p+2})$. Letting $L_1, L_2, and \ L_3$ be the number of legal possible matches for $\mathcal{M}_1, \mathcal{M}_2, \mathcal{M}_3$ respectively. Since the literals in $\mathcal{M}_1$ are variable-distinct from the literals in $\mathcal{M}_2$, we have that the number of legal possible matches, $L$, for the t-3-module $\mathbf{M}$ satisfies

$$L_1 L_2 \leq L \leq 27 L_1 L_2.$$

We now estimate the order of $L_1$. To this end, we consider the probability space $(\Omega, P)$, where $\Omega$ is the set of sequences $(g_1, \cdots, g_p)$ of local fitness functions that possibly match $\mathcal{M}_1$ and $P$ is the uniform probability distribution. Then, the number of legal possible matches is

$$L_1 = |\Omega| \cdot Pr\{\text{a random sample from } \Omega \text{ is legal}\} \tag{4.5}$$

Let $\mathbf{g} = (g_1, \cdots, g_p)$ be a random sample from $\Omega$ and $x_{g_m}$ denote the main variable of the local fitness function $g_m$, then we have

$$Pr\{x_{g_m} = |u_m|\} = Pr\{x_{g_m} = |u_{m+1}|\} = Pr\{x_{g_m} = |z_m|\} = \frac{1}{3},$$

where $|u|$ denotes the variable corresponding to the literal $u$.

Let $B_m, 0 < m \leq p$ be the event that the first $m$ local fitness functions $g_1, \cdots, g_m$ in the possible match $\mathbf{g} = (g_1, \cdots, g_p)$ are mutually distinct. Since in $\mathcal{M}_1$ only consecutive 3-modules share variables, we have

$$B_m = \{(g_1, \cdots, g_m) : \ g_i \neq g_{i+1}, \ 1 \leq i \leq m - 1\}.$$

Let $b_m = Pr\{g_m \neq g_{m-1} \mid B_{m-1}\}, m \geq 2,$ and $b_1 = 1$. Notice that $B_1 = \Omega$. Then, we have

$$\begin{aligned}
Pr\{\mathbf{g} = (g_1, \cdots, g_p) \text{is legal}\} &= Pr\{B_p\} \\
&= Pr\{g_1 \neq g_2, g_2 \neq g_3, \cdots, g_{p-1} \neq g_p\} \\
&= Pr\{B_1\} Pr\{g_2 \neq g_1 \mid B_1\} \cdot Pr\{g_3 \neq g_2 \mid B_2\} \cdots Pr\{g_p \neq g_{p-1} \mid B_{p-1}\} \\
&= b_1 b_2 \cdots b_p
\end{aligned} \tag{4.6}$$

Recalling that $x_{g_m}$ denotes the main variable of the local fitness function $g_m$, we have

$$\begin{aligned}
b_p &= Pr\{g_{p-1} \neq g_p, x_{g_{p-1}} = |u_p| \mid B_{p-1}\} + Pr\{g_{p-1} \neq g_p, x_{g_{p-1}} \neq |u_p| \mid B_{p-1}\} \\
&= Pr\{g_{p-1} \neq g_p \mid B_{p-1}, x_{g_{p-1}} = |u_p|\} \cdot Pr\{x_{g_{p-1}} = |u_p| \mid B_{p-1}\} + \\
&\quad Pr\{g_{p-1} \neq g_p \mid B_{p-1}, x_{g_{p-1}} \neq |u_p|\} \cdot Pr\{x_{g_{p-1}} \neq |u_p| \mid B_{p-1}\} \\
&= \frac{2}{3} a_p + (1 - a_p) \\
&= 1 - \frac{1}{3} a_p, \tag{4.7}
\end{aligned}$$

where $a_p = Pr\{x_{g_{p-1}} = |u_p| \mid B_{p-1}\}$. For $a_p$, we have

$$
\begin{aligned}
a_p &= \frac{Pr\{B_{p-1}, x_{g_{p-1}} = |u_p|\}}{Pr\{B_{p-1}\}} \\
&= \frac{1}{Pr\{B_{p-1}\}} (Pr\{B_{p-1}, x_{g_{p-1}} = |u_p|, x_{g_{p-2}} = |u_{p-1}|\} \\
&\quad + Pr\{B_{p-1}, x_{g_{p-1}} = |u_p|, x_{g_{p-2}} \neq |u_{p-1}|\}) \\
&= \frac{1}{Pr\{B_{p-1}\}} (Pr\{x_{g_{p-1}} = |u_p| \mid B_{p-1}, x_{g_{p-2}} = |u_{p-1}|\} \cdot Pr\{B_{p-1}, x_{g_{p-2}} = |u_{p-1}|\} \\
&\quad + Pr\{x_{g_{p-1}} = |u_p| \mid B_{p-1}, x_{g_{p-2}} \neq |u_{p-1}|\} \cdot Pr\{B_{p-1}, x_{g_{p-2}} \neq |u_{p-1}|\}) \\
&= \frac{1}{Pr\{B_{p-1}\}} \left( \frac{1}{2} Pr\{B_{p-1}, x_{g_{p-2}} = |u_{p-1}|\} + \frac{1}{3} Pr\{B_{p-1}, x_{g_{p-2}} \neq |u_{p-1}|\} \right)
\end{aligned}
\tag{4.8}
$$

The last equation in the above formula is because that given $B_{p-1}$ and $x_{g_{p-2}} = |u_{p-1}|$ (or $x_{g_{p-2}} \neq |u_{p-1}|$), we have two (three, respectively) choices in selecting the local fitness function $g_{p-1}$. Consider the two terms $Pr\{B_{p-1}, x_{g_{p-2}} = |u_{p-1}|\}$ and $Pr\{B_{p-1}, x_{g_{p-2}} \neq |u_{p-1}|\}$ in (4.8), we have

$$
\begin{aligned}
&Pr\{B_{p-1}, x_{g_{p-2}} = |u_{p-1}|\} \\
&= Pr\{g_{p-2} \neq g_{p-1} \mid B_{p-2}, x_{g_{p-2}} = |u_{p-1}|\} \cdot Pr\{B_{p-2}, x_{g_{p-2}} = |u_{p-1}|\} \\
&= \frac{2}{3} \Pr\{x_{g_{p-2}} = |u_{p-1}| \mid B_{p-2}\} \cdot Pr\{B_{p-2}\} \\
&= \frac{2}{3} a_{p-1} \cdot Pr\{B_{p-2}\}
\end{aligned}
\tag{4.9}
$$

and

$$
\begin{aligned}
&Pr\{B_{p-1}, x_{g_{p-2}} \neq |u_{p-1}|\} \\
&= Pr\{g_{p-2} \neq g_{p-1} \mid B_{p-2}, x_{g_{p-2}} \neq |u_{p-1}|\} \cdot Pr\{B_{p-2}, x_{g_{p-2}} \neq |u_{p-1}|\} \\
&= Pr\{x_{g_{p-2}} \neq |u_{p-1}| \mid B_{p-2}\} \cdot Pr\{B_{p-2}\} \\
&= (1 - a_{p-1}) \cdot Pr\{B_{p-2}\}
\end{aligned}
\tag{4.10}
$$

By plugging (4.9) and (4.10) into (4.8), we get

$$
a_p = \frac{Pr\{B_{p-2}\}}{Pr\{B_{p-1}\}} \left( \frac{1}{3} a_{p-1} + \frac{1}{3}(1 - a_{p-1}) \right) = \frac{1}{3 b_{p-1}}.
$$

This, together with (4.7), gives us

$$
b_p = 1 - \frac{1}{9 b_{p-1}}.
\tag{4.11}
$$

It is not difficult to show that the sequence $\{b_p\}$ is decreasing and lower bounded by 0. Letting $\lim\limits_{p} b_p = b$ and taking the limit on both sides, we get

$$
b = 1 - \frac{1}{9b},
\tag{4.12}
$$

and thus, $b = \frac{3 \pm \sqrt{5}}{6}$. In our case, $b = \frac{3+\sqrt{5}}{6}$ since $b_1 = 1$. It follows that $b_p \geq b = \frac{3+\sqrt{5}}{6}$ and thus,

$$b_1 \cdots b_p \geq \left( \frac{3+\sqrt{5}}{6} \right)^p.$$

From (4.5), we know that the number of legal possible matches is greater than

$$3^p \left( \frac{3+\sqrt{5}}{6} \right)^p = \left( \frac{3+\sqrt{5}}{2} \right)^p. \tag{4.13}$$

To prove that the expected number of legal possible matches $L_1$ for $\mathcal{M}_1$ is in $\Theta \left( \left( \frac{3+\sqrt{5}}{2} \right)^p \right)$, let $\alpha_p = b_p - \frac{3+\sqrt{5}}{6} = b_p - b$. From (4.11) and (4.12), we have

$$\alpha_p = b_p - b = \frac{b_{p-1} - b}{9bb_{p-1}} \leq d\alpha_{p-1}, \quad 0 < d < 1,$$

which means that the series $\sum\limits_{m=1}^{p} \alpha_m$ is convergent. It follows that

$$(1 + \frac{\alpha_1}{b}) \cdots (1 + \frac{\alpha_p}{b})$$

converges to a finite positive constant $c$. Therefore,

$$\begin{aligned} b_1 \cdots b_p &= (b + \alpha_1) \cdots (b + \alpha_p) \\ &= b^p \left( 1 + \frac{\alpha_1}{b} \right) \cdots \left( 1 + \frac{\alpha_p}{b} \right) \\ &\leq c \left( \frac{3+\sqrt{5}}{6} \right)^p \end{aligned} \tag{4.14}$$

for sufficient large $p$ and some constant $c$.

Similarly, we can show that the number of legal possible matches $L_2$ for $\mathcal{M}_2$ is in $\Theta \left( \left( \frac{3+\sqrt{5}}{2} \right)^{2p+2} \right)$. Recalling that the number of legal possible matches $L$ for the t-3-module satisfies $L_1 L_2 \leq L \leq 27 L_1 L_2$, the second conclusion follows. $\quad\square$

The following Lemma calculates the probability that a matching local fitness function implies the matched 3-module.

**Lemma 4.3.** *Given a 3-module $x \vee y \vee w$, $x \vee y \vee \bar{w}$, and a local fitness function $g$ such that the main variable $x_g$ of $g$ is one of the three Boolean variables $|x|, |y|, |w|$, let $z = 2 + \alpha$, $0 \leq \alpha \leq 1$ be the parameter in the fixed ratio model $N(n, 2, z)$. Then the probability that $g$ contains the 3-module is*

$$p_0 = \left( \frac{1}{\binom{n-1}{2}} \right) \left( \frac{1}{28}(1-\alpha) + \frac{6}{56}\alpha \right) \tag{4.15}$$

*Proof:* Since $x_g$ is already one of the variables in the 3-module, the probability that the other two variables are also in the 3-module is $\frac{1}{\binom{n-1}{2}}$.

Now, assume that the variables of the local fitness function $g$ are the same as the variables in the 3-module. From the definition of the fixed ratio model, $g$ has two zeros in its fitness value assignment with probability $(1 - \alpha)$, and has three zeros in its fitness assignment with probability $\alpha$. Note that the local fitness function $g$ implies the 3-module $x \vee y \vee w$, $x \vee y \vee \bar{w}$ if and only if

$$g(\bar{x}, \bar{y}, \bar{w}) = 0 \quad \text{and} \quad g(\bar{x}, \bar{y}, w) = 0.$$

From the definition of the fixed ratio model, this happens with the probability

$$\frac{1}{\binom{8}{2}}(1 - \alpha) + \frac{6}{\binom{8}{3}}\alpha$$

The Lemma follows. $\quad \square$

With the above preparation, we can now prove that the average number of t-3-modules contained in $N(n, 2, z)$ tends to infinity.

**Theorem 4.2.** *Let $A_t$ be the number of t-3-modules contained in $N(n, 2, z)$ and $t = \Theta(\ln^2 n)$. Then, if $z = 2 + \alpha > 2.837$,*

$$\lim_{n \to \infty} E\{A_t\} = \infty. \tag{4.16}$$

*Proof:* From Lemma 4.2, there are more than $\left(\frac{3+\sqrt{5}}{2}\right)^t$ legal possible matches for a fixed t-3-module. From Lemma 4.3, we know that each possible legal match $\mathbf{g} = \{g_1, \cdots, g_t\}$ implies the t-3-module with probability $p_0^t$. From the proof of Theorem 10.1 in (Franco & Gelder, 1998), there are

$$2^{t-2} n^{\underline{t-1}}(n - t + 1)^t \tag{4.17}$$

possible t-3-modules, where $n^{\underline{t-1}} = \frac{n!}{(n-t+1)!}$. Let $r = \left(\frac{1}{28}(1 - \alpha) + \frac{6}{56}\alpha\right)$, and write $p_0 = \frac{1}{\binom{n-1}{2}} r$. We have

$$
\begin{aligned}
E\{A_t\} &= \left(\frac{3+\sqrt{5}}{2} p_0\right)^t \cdot 2^{t-2} n^{\underline{t-1}}(n - t + 1)^t \\
&= \left(\frac{3+\sqrt{5}}{2} r\right)^t \cdot 2^{t-2} n^{\underline{t-1}}(n - t + 1)^t \cdot \frac{1}{\binom{n-1}{2}^t} \\
&= \frac{1}{4(n - t + 1)} \left(\frac{3+\sqrt{5}}{2} r\right)^t \cdot \frac{2^t n^{\underline{t}}(n - t + 1)^t}{\binom{n}{2}^t} \left(\frac{\binom{n}{2}}{\binom{n-1}{2}}\right)^t \\
&= \frac{1}{4(n - t + 1)} \left(\frac{3+\sqrt{5}}{2} r\right)^t \cdot \frac{4^t n^{\underline{t}}(n - t + 1)^t}{(n(n - 1))^t} \left(\frac{n}{n - 2}\right)^t \\
&= \frac{1}{4n}(2(3 + \sqrt{5})r)^t \left(1 - O\left(\frac{t^2}{n}\right)\right),
\end{aligned}
\tag{4.18}
$$

where the fourth equation in (4.18) is due to the fact that for any positive integer n and q such that $q < \frac{n}{2}$, we have $n^q e^{-q^2/2n} \leq n\underline{q} \leq n^q$. It follows that $\lim_{n \to \infty} E\{A_t\} = \infty$ if

$$2(3 + \sqrt{5})r > 1. \tag{4.19}$$

Solving the inequality (4.19) gives us $\alpha > 0.837$, that is, $z = 2 + \alpha > 2.837$. This proves Theorem 4.2. $\square$

Based on the Chebychev's inequality, to prove that $N(n, 2, z)$ contains t-3-modules with probability 1, we need to show that the variance of $A_t$, the number of contained t-3-modules, is $o(E\{A_t\})$. For this purpose, we follow Franco and Gelder's approach (Lemma 4.1, Franco & Gelder, 1998) to apply the second moment method (Alon & Spencer, 1992):

**Lemma 4.4.** *(Alon & Spencer, 1992, Ch. 4.3 Cor 3.5) Given a random structure(e.g., a random CNF formula), let $W$ be the set of substructures under consideration, $A(w)$ be the set of substructures sharing some clauses with $w \in W$. Let $I_w = 1$ when $w$ is in the random structure and 0 otherwise. If*

*(1) elements of $W$ are symmetric;*

*(2) $\mu = E\{\sum_{w \in W} I_w\} \to \infty$; and*

*(3) $\sum_{\bar{w} \in A(w)} Pr(\bar{w} \mid w) = o(\mu)$, for each $w \in W$,*

*then as $n \to \infty$, the probability that the random structure contains a substructure tends to 1.*

To use the above Lemma to study the 2-SAT sub-problem in NK landscapes, we view the random structure to be a random instance of $N(n, 2, z)$, and $W$ to be the set of all t-3-modules which are symmetric by their definitions(Sections 5 and 10, Franco & Gelder, 1998).

**Theorem 4.3.** *If $z = 2 + \alpha > 2.837$, then $N(n, 2, z)$ is asymptotically insoluble with probability 1.*

**Proof:** Let $A_t$ be the number of t-3-modules implied by $N(n, 2, z)$ and $t = O(\ln^2 n)$. Theorem 4.2 shows that $\lim_{n \to \infty} E\{A_t\} = \infty$. By Lemma 4.4, it is enough to show that for each $w \in W$,

$$\sum_{\bar{w} \in A(w)} Pr(\bar{w} \mid w) = o(E\{A_t\}), \tag{4.20}$$

where $Pr(\bar{w} \mid w)$ is the conditional probability that $N(n, 2, z)$ implies the t-3-module $\bar{w}$ given that it implies $w$, and $A(w)$ is the set of all t-3-modules sharing some clauses with $w$.

Suppose that $\bar{w}$ shares $Q, 1 \leq Q \leq 2t$ clauses with $w$, and that these $Q$ clauses are distributed among $q$ 3-modules. Further, let $q_1$ be the number of 3-modules whose two clauses are both shared and $q_2 = q - q_1$ the number of 3-modules that only has one clause shared.

Let $T_1$ be a 3-module in $\bar{w}$ that shares exactly one clause with a 3-module $T_2$ in $w$. We claim that the conditional probability that $T_1$ is implied by $N(n, 2, z)$ given that $w$ is implied by $N(n, 2, z)$, is

$$\frac{1}{6}\alpha + O(\frac{1}{n}). \tag{4.21}$$

Without loss of generality, assume that $T_2 = \{x \vee y \vee u, x \vee y \vee \bar{u}\}$ and $T_1 = \{x \vee y \vee u, \bar{x} \vee y \vee \bar{u}\}$. Since $w$ is implied by $N(n, 2, z)$, there is a local fitness function $g = g(|x|, |y|, |u|)$ that implies $T_2$. The conditional probability that $T_1$ is implied, is less than or equal to $P_1 + P_2$ where $P_1$ is the conditional probability that $g$ also implies the clause $\bar{x} \vee y \vee \bar{u}$ given that $g$ implies $T_2$, and $P_2$ is the conditional probability that the clause $\bar{x} \vee y \vee \bar{u}$ is implied by other local fitness functions. By the definition of $N(n, 2, z)$, we have that $P_1 = \frac{1}{6}\alpha$. Since a local fitness function implies $\bar{x} \vee y \vee \bar{u}$ only if it has the same variables with $g = g(|x|, |y|, |u|)$, we have that $P_2 = O(\frac{1}{n})$. The claim is proved. It follows that, for sufficiently large $n$,

$$Pr\{\bar{w} \mid w\} \le c \left(\frac{3 + \sqrt{5}}{2} p_0\right)^{t-q} \cdot 1^{q_1} \cdot \left(\frac{1}{6}\alpha\right)^{q_2} \tag{4.22}$$

where $p_0$ is defined in Lemma 4.3 and $c$ is a fixed constant.

Let $A_{Q,q,q_2}(w)$ be the set of t-3-modules that share $Q$ clauses with $w$ such that these $Q$ clauses are distributed over $q$ different 3-modules. As before, $q_1$ is the number of 3-modules whose two clauses are both shared and $q_2 = q - q_1$ the number of 3-modules that only has one clause shared. We claim that

$$|A_{Q,q,q_2}(w)| = |A_{2q,q,0}(w)|6^{q_2}. \tag{4.23}$$

where $A_{2q,q,0}(w)$ is the set of t-3-modules that share all the $2q$ clauses in the $q$ 3-modules with $w$. Let $\overline{M} = \{\overline{M_1}, \cdots, \overline{M_t}\}$ be a t-3-module in which all the clauses $\overline{M_i}, 1 \le i \le q$ are shared with $w$. Let $M = \{M_1, \cdots, M_t\}$ be a t-3-module in which all the clauses in $M_i, 1 \le i \le q_1$ are shared and each of the 3-modules $M_i, q_1 + 1 \le i \le q_1 + q_2$ has only one clause shared. Since for each of the $q_2$ 3-modules, we have 6 ways to choose the non-shared clauses, there are $6^{q_2}$ such t-3-modules $M$ in $A_{Q,q,q_2}(w)$ that correspond to one t-3-module $\overline{M}$ in $A_{2q,q,0}$. The claims follow. From formula (55) and (56) in (Franco & Gelder, 1998) and (4.23), it follows that

$$|A_{Q,q,q_2}(w)| < \begin{cases} \frac{O(t)}{n^2} 2^{t-q} n^{2(t-q)} 6^{q_2} & , q \le p + 1, \\ \frac{O(1)}{n} 2^{t-q} n^{2(t-q)} 6^{q_2} & , q > p + 1. \end{cases} \tag{4.24}$$

Let $r = \left(\frac{1}{28}(1 - \alpha) + \frac{6}{56}\alpha\right)$, and write $p_0 = \frac{1}{\binom{n-1}{2}} r$. Then, we have

$$\begin{aligned} &|A_{Q,q,q_2}(w)|Pr\{\bar{w} \mid w\} \\ &\le \frac{O(t)}{n^2} 2^{t-q} n^{2(t-q)} 6^{q_2} \left(\frac{3 + \sqrt{5}}{2} p_0\right)^{t-q} \left(\frac{1}{6}\alpha\right)^{q_2} \\ &\le \frac{O(t)}{n^2} 2^{t-q} n^{2(t-q)} \left(\frac{3 + \sqrt{5}}{2} r\right)^{t-q} \frac{1}{\binom{n-1}{2}^{t-q}} \\ &\le \frac{O(t)}{n} \frac{1}{4n} (2(3 + \sqrt{5})r)^{t-q} \\ &\le \frac{O(t)}{n} E\{A_t\} (2(3 + \sqrt{5})r))^{-q}, \quad q \le p + 3 \end{aligned} \tag{4.25}$$

and

$$|A_{Q,q,q_2}(w)|Pr\{\bar{w} \mid w\} \leq O(1)E\{A_t\}(2(3+\sqrt{5})r))^{-q}, \quad q > p+3. \qquad (4.26)$$

Therefore,

$$\sum_{\bar{w} \in A(w)} Pr(\bar{w} \mid w) = \sum_{Q,q,q_2} |A_{Q,q,q_2}(w)|Pr\{\bar{w} \mid w\}$$

$$= \sum_{Q=1}^{t} \sum_{q \leq p+3} \sum_{q_2} \frac{O(t)}{n}E\{A_t\}(2(3+\sqrt{5})r))^{-q} + \sum_{Q=1}^{t} \sum_{q>p+3} \sum_{q_2} O(1)E\{A_t\}(2(3+\sqrt{5})r))^{-q}. \qquad (4.27)$$

Since $2(3+\sqrt{5})r) > 1$ for $z > 2.837$, we have

$$\sum_{\bar{w} \in A(w)} Pr(\bar{w} \mid w) \leq \frac{O(t^4)}{n}E\{A_t\} + t^3 E\{A_t\}(4r)^{-(p+3)} \qquad (4.28)$$

$$= o(E\{A_t\}).$$

This completes the proof of Theorem 4.3. □

## 5. Experiments

Our study of the threshold phenomena in NK landscapes started with an experimental investigation. Many of the theoretical results in the previous section are motivated by the observations made in our experiments. In this section, we describe the approach and methods we used in the experimental study, and report the results and observations we have made.

In our experiments, an instance of the NK landscape decision problem is converted to an equivalent 3-SAT problem, and then the 3-SAT problem is solved using Roberto's relsat—an enhanced version of the famous Davis-Putnam algorithm for SAT problems implemented in $C^{++}$. The source code of relsat can be found at http://www.cs.ubc.ca/ hoos/SATLIB/.

In the experiments, we generated random instances of the NK landscape decision problem from the random model $N(n, 2, z)$. As a result, the equivalent SAT problem for each random NK landscape instance is a 3-SAT problem with $n$ variables and (on average) $zn$ clauses. By definition, the parameter $z$ is between 0 and 8. For $z \leq 1$, the 3-SAT instance can be solved easily by setting the literals that correspond to the main variables of the local fitness function to true. As $z$ increases, we get more and more clauses and the 3-SAT problem becomes more and more constrained. The aims of the experiments are three-fold:(1)Investigating if there exists a threshold phenomenon in the random NK landscape model; (2) Locating the threshold of the parameter $z$; and (3)Determining if there are any hard instances around the threshold.

### 5.1 Experiments on the Fixed Ratio Model

In this part of the experiments, we generate 100 random instances of $N(n, 2, z)$ for each of the parameters $n = 2^9 \cdots 2^{16}$ and $z = 2.71, 2.72, \cdots, 3.00$. These instances are then converted to 3-SAT instances and solved by *relsat*. Figure 1 shows the fraction of insoluble instances as a function of the parameter $z$. It can be seen that there exists a threshold phenomenon and the threshold is around 2.83. This shows that our upper bound $z = 2.837$ is very tight.

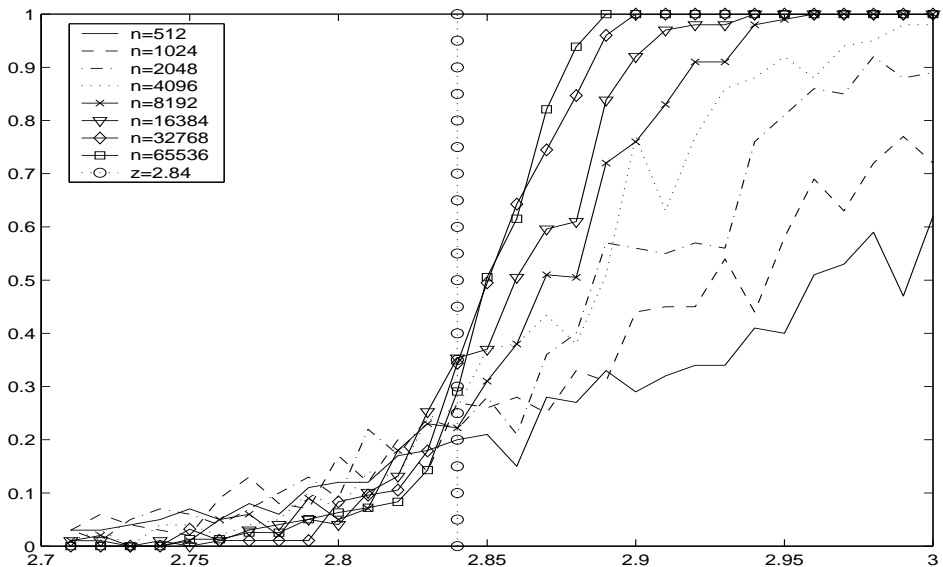

Figure 1: Fractions of insoluble instances(Y-axis) as a function of $z$ (X-axis).

In Figure 2, we plot the square root of the average search cost as a function of the parameter $n$. The figure indicates that the average search is in $O(n^2)$ for any parameter $z$. We have also observed that more than 99 percent of the insoluble instances are solved quickly in the preprocessing stage of relsat. This indicates that there must be some "small" structures that make the instances insoluble. More detailed experimental results can be found in Gao's thesis (Gao, 2001).

## 5.2 Experiments on the 2-SAT sub-Problem

This is the part of the experiments that motivated our theoretical analyses in Section 4.2. The idea can be explained as follows. Let

$$f(x) = \sum_{i=1}^{n} f_i(x_i, \Pi(x_i))$$

be an instance of the decision problem of NK landscape and

$$\varphi = C_1 \bigwedge C_2 \cdots \bigwedge C_n$$

the equivalent 3-SAT problem where $C_i$ is the set of 3-clauses equivalent to the local fitness function $f_i$. For each $i$, there is a set of 2-clauses $D_i$(possibly empty) implied by $C_i$. For

327

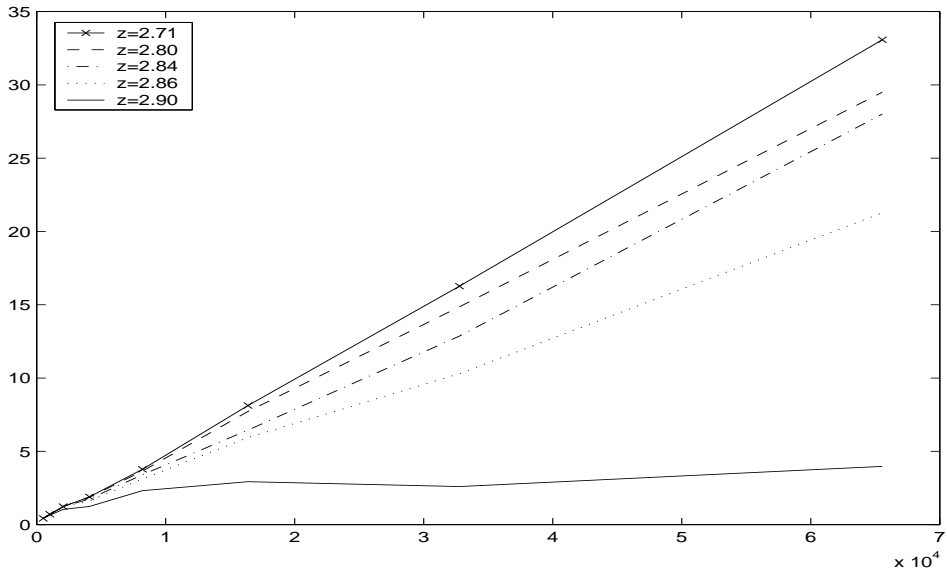

Figure 2: Square root of the average search cost (Y-axis, in seconds) as a function of $n$ (X-axis).

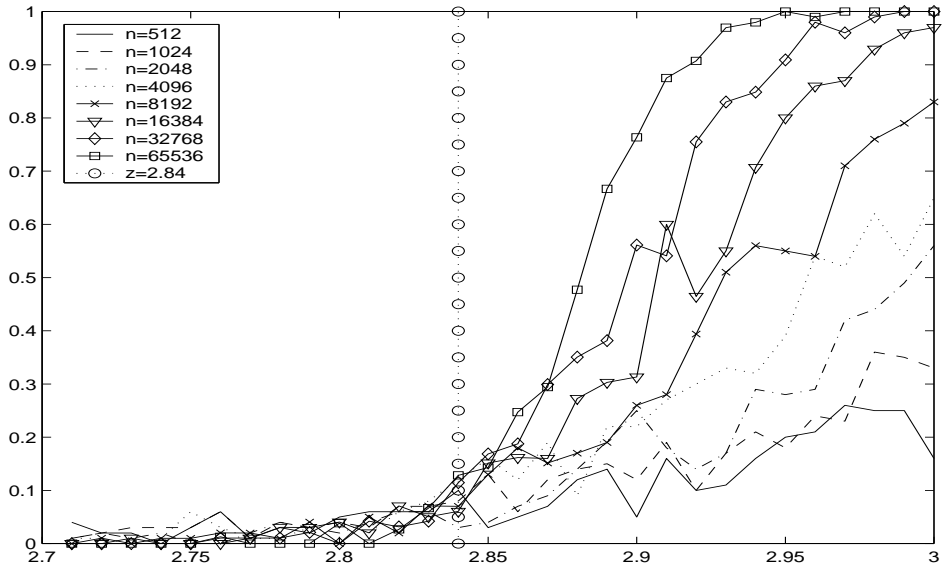

Figure 3: Fractions of insoluble instances(Y-axis) as a function of $z$ (X-axis) for 2-SAT sub-problems.

example, if $C_i$ has three 3-clauses $((x, y, z), (x, \bar{y}, z), (x, y, \bar{z}))$, then the set of 2-clauses $D_i$ would be $((x, z), (x, y))$. The conjunction of $D_i$, denoted by $\bar{\varphi}$, is a 2-SAT problem. It is obvious that the original 3-SAT problem $\varphi$ is satisfiable only if the 2-SAT sub-problem $\bar{\varphi}$ is satisfiable. In the experiment, we generate instances of the NK landscape $N(n, 2, z)$, convert them to the equivalent 3-SAT problems, and extract the 2-SAT sub-problems. These 2-SAT problems are then solved by the relsat solver. If the 2-SAT problem is unsatisfiable, then the original NK landscape instance is also insoluble.

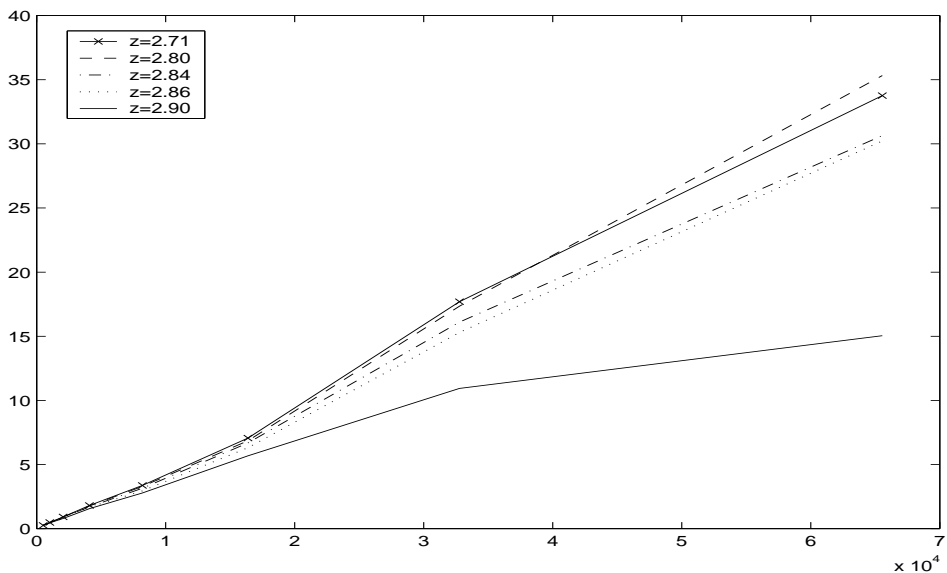

Figure 4: Square root of the average search cost (Y-axis, in seconds) as a function of $n$ (X-axis) for 2-SAT sub-problems.

The experimental settings are the same as those in the experiment on the original problem. The results are shown in Figures 3-4, in parallel to the Figures 1-2 of the results on the original 3-SAT problems in Section 5.1. We see that the patterns of insoluble fractions and search cost are similar to those we found in the original 3-SAT problems. There is a soluble-insoluble phase transition occurring around 2.83, but the fraction of unsatisfiable instances is lower than the fraction in the original 3-SAT problems.

We also observed that the average search cost for the 2-SAT sub-problems remains the same as that for the original 3-SAT problems. This tells us that the difficulty of solving a soluble instance of NK landscape is almost the same as that of solving a 2-SAT problem, and hence is easy. Therefore, on average the NK landscape $N(n, 2, z)$ is also easy at parameters below the threshold where almost all of the instances are soluble.

## 6. Implications and Conclusions

One of the questions that arises about this work is its implications to the design and analysis of genetic algorithms. NK landscapes were initially conceived as simplified models of evolutionary landscapes which could be tuned with respect to ruggedness and epistatic interactions (Kauffman, 1989). In the study of genetic algorithms, NK landscape models have been used as a prototype and benchmark in the analysis of the performance of different genetic operators and the effects of different encoding methods on the algorithm's performance (Altenberg, 1997; Hordijk, 1997; Jones, 1995). Kauffman (1993) points out that the parameters that primarily affect a number of ruggedness measures are $n$ and $k$.

Nevertheless, the fact that for $k \geq 2$ the discrete NK landscape is NP-complete (Wright et al., 2000) when the neighbors are arbitrarily chosen could be construed as implying that random landscapes with fixed $k$ are in practice hard.

The results in this paper should serve as a cautionary note that this may not be the case. Our analyses show that for fixed $k$ the uniform probability model is trivially solvable as the problem size tends to infinity. For the fixed ratio model, we have derived two upper bounds for the threshold of the solubility phase transition, and proved that the problem with the control parameter above the upper bounds can be solved in polynomial time with probability asymptotic to 1 due to the existence of easy sub-problems such as 2-SAT. A series of experiments has also been conducted to investigate the hardness of the problem with the control parameters around and below the threshold. From the experiments, we have observed that the problem is also easy around and below the threshold.

Our proofs hold only for the decision version of the problem where the component functions are discrete on $\{0, 1\}$. The proofs are obtained by noticing that the clustering of functions, or clauses, on selected subsets of variables implies that the overall problem is decomposable into independent subproblems, or that the problem contains small substructures that identify the solution. The subproblems are the components of the connection graph defined in Section 3 and the 2-SAT sub-problems studied in Section 4.2. It is currently unclear to us to what extent our analysis can be extended to the optimization version of the NK model, and we would like to study this problem further in the future.

In response to the question 'what are the implications for GAs?' we suggest the following speculative line of enquiry. For the discrete model we use, the soluble instances are readily solved by a standard algorithmic approach based on recognizing the components of the connection graph. (This should not be a surprise for us as it has been pointed out by Heckendorn, Rana, and Whitley (1999) that 'Even relatively old algorithms such as Davis-Putnam which are deterministic and exact are orders of magnitude faster than GAs'.) [1] A similar connectivity can be developed for real valued distributions, for example by capping the minimum value which we allow a sub-function to take. We can speculate that the clustering imposed by fixed values of $k$ would also generate localized structures when real values are applied and when considering optimization instead of decision, but perhaps with fuzzy boundaries. In fact, this observation is just the flip side of limited epistasis. Genetic algorithms, or their variants such as the probabilistic model-building algorithms (Larranaga & Lozano, 2001), designed to mimic natural evolution, are supposed to take advantage of this situation. So, to the extent that NK landscapes are an accurate reflection of the features exploited by evolutionary algorithms, we pose the following question. Is it possible to identify these fuzzy components if they exist, and in doing so design an algorithm that exploits the same landscape features that the evolutionary algorithms do, but far more efficiently, as we have done for the uniform discrete decision problem?

---

1. We thanks an anonymous referee for pointing out to us the work of Heckendorn, et al. (Heckendorn et al., 1999)

These landscapes were designed with the intent of studying limited interactions, and our results can also be seen as a confirmation that indeed limited epistasis leads to easier problems. In another domain, that of the more traditional research into search and optimization, there is a need for test bed problems with real world connections which are tunable with respect to difficulty. NK landscapes might have been such a domain for generating 3-SAT instances. It is disappointing that for restricted $k$ the instances generated are easy with high probability.

## Acknowledgments

This research supported in part by Natural Sciences and Engineering Research Council Grant No. OGP8053. We thank the anonymous reviewers for their comments.